\documentclass[conference]{IEEEtran}
\IEEEoverridecommandlockouts
\usepackage{cite}
\usepackage{amsmath,amssymb,amsfonts}
\usepackage{algorithmic}
\usepackage{graphicx}
\usepackage{textcomp}
\usepackage{xcolor}
\def\BibTeX{{\rm B\kern-.05em{\sc i\kern-.025em b}\kern-.08em
    T\kern-.1667em\lower.7ex\hbox{E}\kern-.125emX}}
    
\begin{document}

\title{Multi class activity classification in videos using Motion History Image generation
\thanks{https://github.com/sengopal/cv-activity-classification}
}

\author{Senthilkumar Gopal \\
\IEEEauthorblockA{\textit{College of Computing} \\
\textit{Georgia Institute of Technology}\\
\texttt{sengopal@gatech.edu} \\
}
}

\maketitle

\begin{abstract}
Human action recognition has been a topic of interest across multiple fields ranging from security to entertainment systems. Tracking the motion and identifying the action being performed on a real time basis is necessary for critical security systems. In entertainment, especially gaming, the need for immediate responses for actions and gestures are paramount for the success of that system. We show that Motion History image has been a well established framework to capture the temporal and activity information in multi dimensional detail enabling various usecases including classification. We utilize MHI to produce sample data to train a classifier and demonstrate its effectiveness for action classification across six different activities in a single multi-action video as illustrated in Figure \ref{fig:mhi-multi-action-labels}. We analyze the classifier performance and identify usecases where MHI struggles to generate the appropriate activity image and discuss mechanisms and future work to overcome those limitations.

\end{abstract}


\section{Introduction}
Prior work in this domain had relied on location of joins in each frame to form a 3D model of the human body to determine motion \cite{samanta2016data} or \cite{rehg1995model}, or using window tracking methods \citenum{davis1996appearance} to represent human postures and track their symmetrical movements. 

Motion History Image (MHI) is a view based method to represent motion across time as per \cite{davis1999recognizing}. This is widely used for activity recognition and other movement based inferences. The following utilizes the MHI approach presented in \cite{bobick1997action} which proposes an appearance-based recognition strategy. While their first approach \cite{bobick1997action} describes an approach to collapse the region-based motion into a co-efficient vector, the second approach converts the binary motion image to a motion history image (MHI) which acts as a pixel intensity based function of the motion history in a particular location, with brighter pixels representing recent motion. Six different activity classifications were attempted using MHI and the below captures the methods for deriving the MHI and its representational moments and the use of different classifiers for activity recognition.

% All the code used to perform the experiments and results are published for reference \footnote{https://github.com/sengopal/cv-activity-classification}

\begin{figure}[hbt!]
    \centering
    \includegraphics[width=1\linewidth]{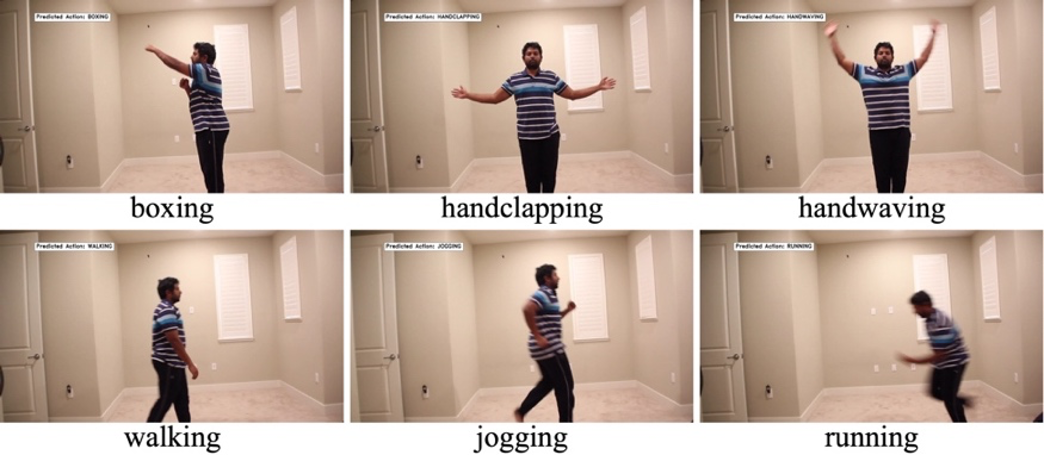}
    \caption{Multi Action video frames with their predicted action labels}
    \label{fig:mhi-multi-action-labels}
\end{figure}

\section{Related Work}

\subsection{Temporal Templates}

The inspiration for using ``patterns of motion'' to track an action was based on the comic-strip panels which attempt to capture character motion in a single static frame. A character flying or punching is represented as a panning of a moving image which intuitively represents the action being performed. Such a static representation forms the basis of temporal templates.

\subsection{Binary Images and background subtraction}
In order to capture the motion over time, the person performing the motion and the binary motion signal has to be extracted over time. 
To compute this binary motion, cumulative frame differences are calculated using a threshold of $\theta$. Also to ensure noise reduction, the frame images are pre-processed with a Gaussian filter of kernel $(3,3)$ to eliminate any image compression artifacts. An example set of binary images from the provided dataset is represented below.

\begin{figure}[hbt!]
    \centering
    \includegraphics[width=1\linewidth]{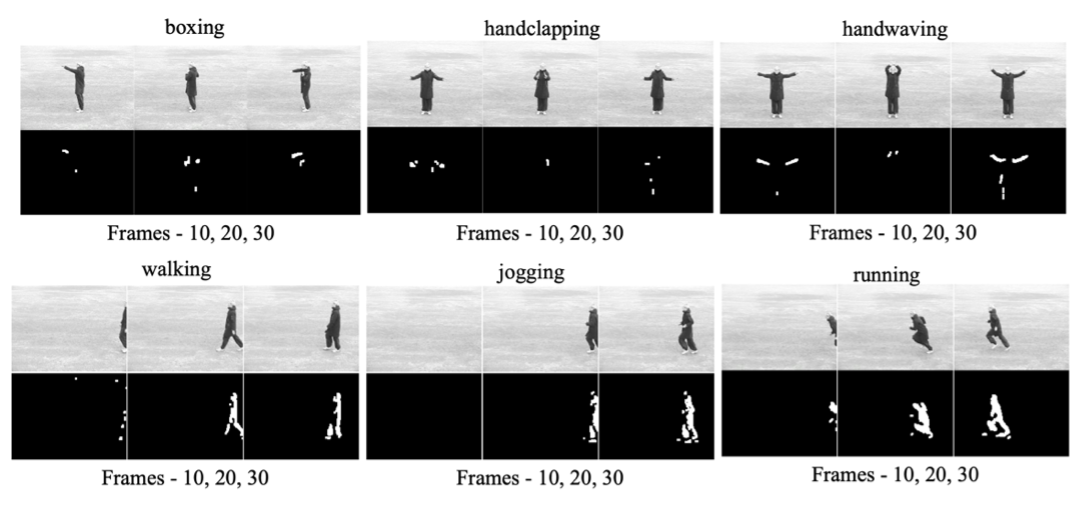}
    \caption{Illustrative binary difference images for various actions}
    \label{fig:training-samples}
\end{figure}

To further prevent noise artifacts within the binary images, the morphological OPEN operator \cite{vincent1993morphological} and \cite{Maragos1986} was used where the reconstruction eliminates the noisy pixels within the binary signal and provides a consistent, smooth signal.

The cumulative binary images below provide a better visualization of the different motion signals that are illustrative of a particular action in a 2D space. These Motion-Energy Images (MEI) highlight regions in the image where any form of motion is present and the shape of the region can be uniquely used to determine the action and its viewing conditions.

\begin{figure}[hbt!]
    \centering
    \includegraphics[width=1\linewidth]{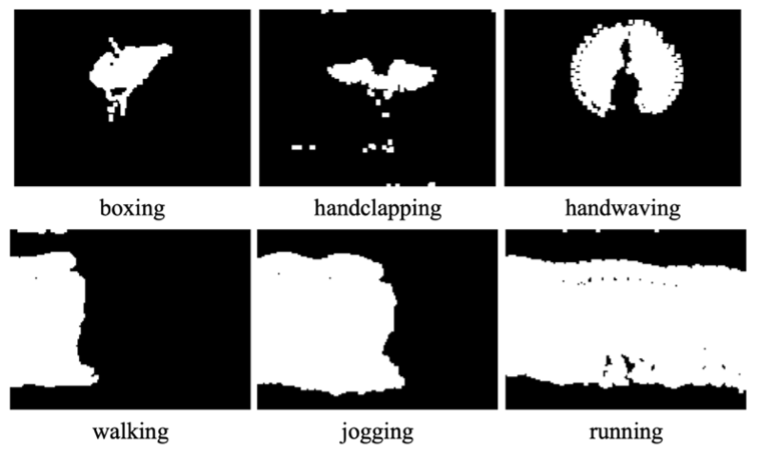}
    \caption{Cumulative binary images (Motion Energy Response) for various actions}
    \label{fig:motion-energy-train}
\end{figure}

\subsection{Motion-history images}
The set of binary images over time t, represent the direction and speed of the motion [how] as well as the location of motion [where] as well described in \cite{davis1996appearance}. Motion history image (MHI) combines this information into one static image similar to earlier attempts of representing motion such as \cite{manning1998understanding} which acted as the inspiration for MHI. As elaborated in \cite{manning1998understanding}, artists represent motion in a single frame via \textit{streaking}, where the motion region in the image is blurred. The result is analogous to a picture taken of an object moving faster than the camera's shutter speed, which is termed as \textit{panning}. The recent location of motion is represented brighter than the remaining motion region, where this \textit{fading} of intensity indicating the object's motion path. Using this definition, MHI is represented as a pixel intensity function, where brighter values correspond to most recent motion.

\begin{equation}
H_{\tau}(x, y, t) = 
\begin{cases} 
\tau & \text{if } D'(x, y, t)=1 \\
\max(0, H_{\tau}(x, y, t-1)-1) & \text{otherwise}
\end{cases}
\end{equation}

The advantage of using a MHI over the cumulative binary image (MEI) is the ability to retain direction of the motion and the ability to distinguishing between forward and reverse performance of an action \cite{davis1996appearance}. A sample set of such MHIs for the dataset being investigated is provided below as reference.

\begin{figure}[hbt!]
    \centering
    \includegraphics[width=1\linewidth]{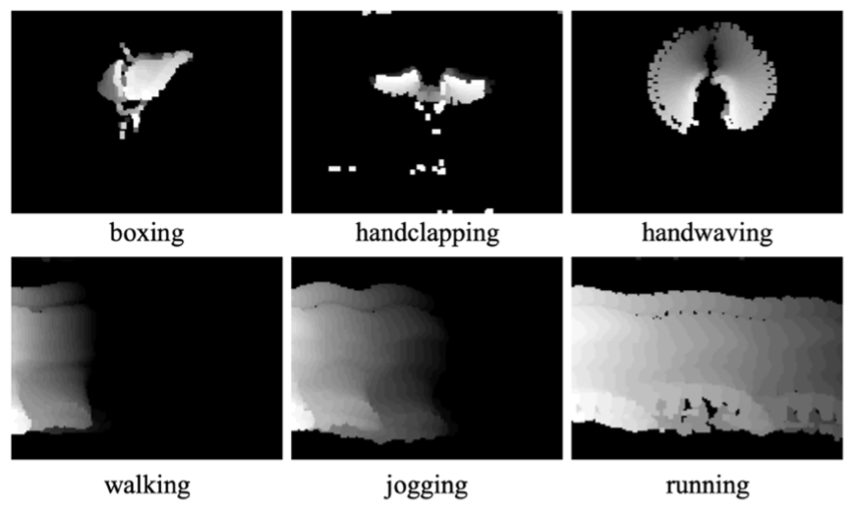}
    \caption{Motion History Image (MHI) for various actions}
    \label{fig:motion-history-train}
\end{figure}

As described earlier, brighter pixels represent most recent motion while the greyer pixels indicate earlier motion regions. So for walking, the grayer pixels are very less as they have decayed over a longer period of time, while for jogging they is lesser decay while for running the decay is the least.

\subsection{Moments}
Continuing with the algorithm defined in \cite{Davis1997RepresentationAM}, once a set of MEIs and MHIs for each action sequence has been generated, unique statistical descriptions of these images can be computed using moment-based features. This paper utilizes the seven Hu moments \cite{hu62visual} which are known to provide reasonable shape discrimination in a translation, rotation and scale invariant manner to accommodate the dataset that is being used. The first seven parameters ${<v1,v2,v3,v4,v5,v6,v7>}$ are the Hu moments. We also add the third order independent moment invariant as determined in \cite{Flusser2000} to ensure independence and completeness.

So, the raw moments for the grayscale image is represented by:

\begin{equation}
M_{ij} = \sum_x \sum_y x^i y^j I(x, y)
\end{equation}

Given this definition, the area for binary images is represented by $M_{00}$ and the centroid definition is:

\begin{equation}
\text{Centroid: } \{\bar{x}, \bar{y}\} = \left\{ \frac{M_{10}}{M_{00}}, \frac{M_{01}}{M_{00}} \right\}
\end{equation}

The central moments can then be calculated for a grayscale image using the centroid moments as 

\begin{equation}
\mu_{pq} = \sum_x \sum_y (x - \bar{x})^p (y - \bar{y})^q f(x, y)
\end{equation}

To ensure scale invariance is also factored in, both central moments (as illustrated above) and scale invariant moments are calculated as:

\begin{equation}
\nu_{pq} = \frac{\mu_{pq}}{\mu_{00}^{(1 + \frac{p+q}{2})}}
\end{equation}

These moments (\textit{central and scale-invariant}) would be used to uniquely identify the MHI+MEI temporal templates for action recognition.

\subsection{Importance of Tau}
As described in \cite{davis1996appearance}, the temporal segmentation of actions is an important aspect of MHI computation. The time-window length $\tau$ defines the extent of action being recorded in the MHI and MEI. For the current report, due to the limited time duration, a fixed $\tau$ time window was used as \textbf{300} for both training and testing phases.

\subsection{Image processing Pipeline}
\begin{figure}[hbt!]
    \centering
    \includegraphics[width=1.0\linewidth]{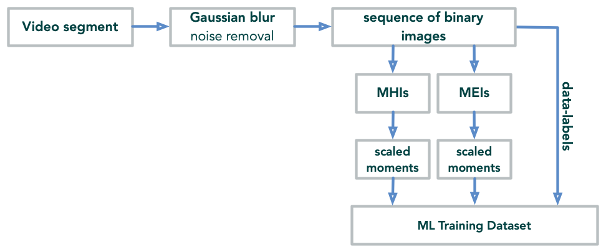}
    \caption{Image processing pipeline to generate moments}
    \label{fig:pipeline}
\end{figure}

Putting all of the above together, the image processing pipeline given an input video or a sequence of frame images is illustrated in Figure \ref{fig:pipeline}. The output of the pipeline will be then feed into a machine learning algorithm capable of classification along with known data labels to complete the training using the dataset provided.

\section{Classification}
Once the input videos have been converted into dataset as illustrated above, two types of classifiers were utilized for training and then action recognition by splitting the provided dataset randomly into training/validation/testing at 50/25/25 split. Only the dataset with \textit{static homogeneous background} was used for training and action recognition.

\subsection{K Nearest Neighbors Classifier}

\begin{figure}[hbt!]
    \centering
    \includegraphics[width=1\linewidth]{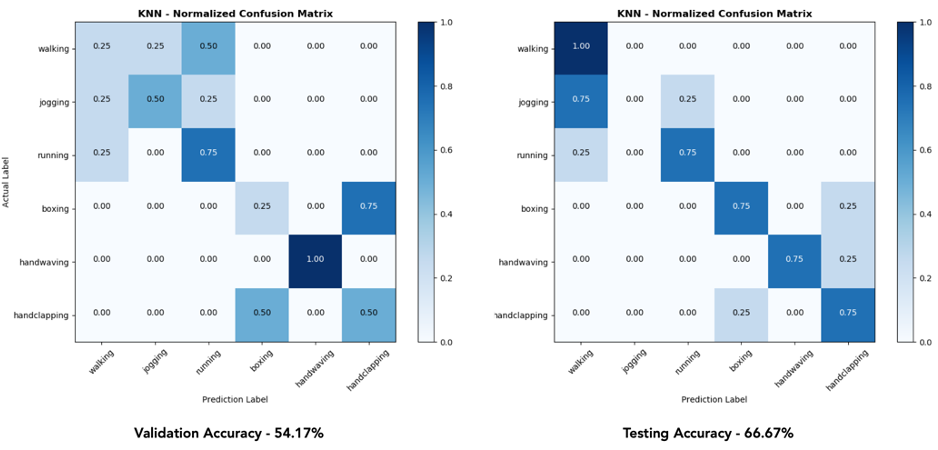}
    \caption{KNN Confusion Matrix for Training and Testing}
    \label{fig:knn-conf}
\end{figure}

K-Nearest Neighbors (KNN) is one of the well known classification algorithms for supervised learning. Based on the training data, the classifier chooses the closest matching data set based on Euclidean distance from the provided sample feature set. Using KNN provided by the OpenCV libraries, the classification accuracy and confusion matrix are provided in Figure \ref{fig:knn-conf}.

The KNN classifier performed poorly over a wide range of $\tau$ with no appreciable gains in any particular range. The KNN classifier really suffered with the boundaries for walking and jogging due to the closeness of the MEI/MHI being generated for them. The constant $\tau$ across various dataset means that the sequence window is exactly the same for different groups of people performing the same action but at a varying pace.

\subsection{Multi-layer Perceptron Classifier}
\begin{figure}[hbt!]
    \centering
    \includegraphics[width=1\linewidth]{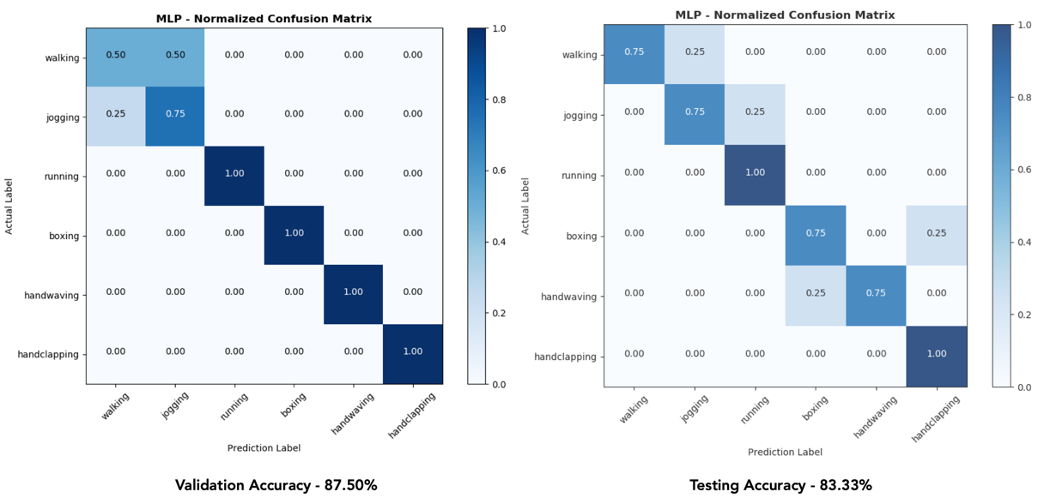}
    \caption{MLP Confusion Matrix for Training and Testing}
    \label{fig:mlp-conf}
\end{figure}

Due to the poor results of KNN, another supervised learning algorithm called Multi-layer Perceptron (MLP) Classifier was used instead. This is a more advanced algorithm which uses neural network like structure to train using hidden layers which captures the subtle differences of the training data and help garner better accuracy. This algorithm uses \cite{backprop} Back-Propagation which continuously adjusts the weights of the network connections to minimize the training/testing output error. Due to these adjustments, the hidden layers \textit{learn} to represent important and subtle differences and aspects while the regular interactions are characterized by the interactions between these units. The MLP has appreciable gains and better results for the same dataset that was used for the KNN classifier. The confusion matrix and the validation/testing results are provided in Figure \ref{fig:mlp-conf}.

\section{Results}
\begin{figure}[hbt!]
    \centering
    \includegraphics[width=1\linewidth]{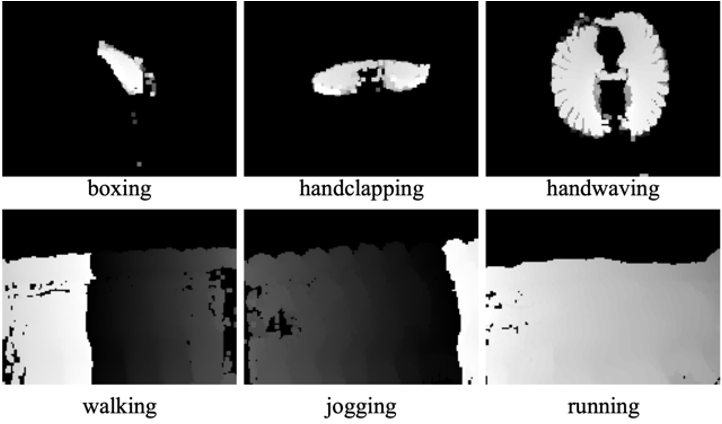}
    \caption{Motion history images for the actions in the multi-action video}
    \label{fig:mhi-multi-action}
\end{figure}

To demonstrate the capability of the developed algorithm, a multi action video was recorded with the same subject performing all of the trained actions in sequence. Then using the image processing pipeline described in Figure \ref{fig:pipeline} earlier, the MHI/MEI images and their Hu moments were computed and then passed to the trained MLP Classifier for recognition which was able to successfully identify all the actions being performed. Some of the motion history image samples along with their training labels are provided in Figure \ref{fig:mhi-multi-action}, while the results on how the labels are applied on the multi-action video were earlier discussed in Figure \ref{fig:mhi-multi-action-labels}.

\section{Analysis}
\subsection{why the method works on some images and not on others}

As part of the prediction for multi action video two other samples were executed which did not work well. One of the samples failed to generate the correct MHIs due to the presence of \textbf{a shadow of the person}, creating a secondary MHI image, corrupting the moments and failing to get the accurate prediction. This brings out the need for better image pre-processing, thresholding and cleanup in general before classification/recognition can be attempted.

\begin{figure}[hbt!]
    \centering
    \includegraphics[width=0.5\linewidth]{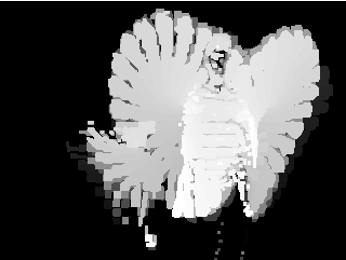}
    \caption{Handwaving MHI with a shadow artifact generated secondary MHI}
    \label{fig:mhi_with_shadow}
\end{figure}

The other major cause for incorrect predictions was the variance in the time-window length $\tau$ which needs to be more tuned for each specific set of video(s). The boundary conditions for the MHI generated for jogging was very close to the running and walking MHIs due to their similarities of motion action. One way to help the classifier become more robust would be to introduce the time for $\tau$ as a factor which in turn can help the classifier be \textbf{$\tau$-invariant}.

\section{Future Work}
One of the more advanced usages of MHI, is the utilization of Motion History Volumes (MHV) where instead of the 2D representation of the motion over time, video recordings of actions are defined as 3D patterns in image-space, and in time, resulting from the perspective projection of the world action onto the image plane at each time instant \cite{weinland2006free}. This has better classification results than the above mentioned experiment, by simply using Euclidean and Mahalanobis distance without any complex classifiers.

Another more recent methodology that was introduced is the building of three-dimensional motion history images computed as the eigenvector of body movements \cite{chang2015research}. After getting the three-dimensional motion history image, this method projects it in the $XY$, $YZ$ and $XZ$ planes. This projection simplifies the invariant moment’s calculation process. Furthermore, this method uses extreme learning machine (ELM) for SLFN which does not adjust most of the parameters of the artificial neural network repeatedly, such as the input weights and the hidden layer biases. Contrary to the traditional artificial neural network, this randomly assigned approach reduces huge amounts of the calculation and increases the artificial neural network’s training efficiency. This method performs exceptionally well in different lighting, background and occlusion conditions \cite{chang2015research}.

\subsection{Further Improvements}
The current experiment was conducted on pre-determined frame sequences. The algorithm can be improved to compute scores for different actions over a sliding window to provide more real-time recognition. The three-dimensional MHI system described in \cite{chang2015research} with ELM would vastly improve the current classifier system with a wider feature set and classification capability. Another improvement would be to have a method of subject isolation to help track their actions better in case of multiple actors being present in the video segment. Lastly, most of the inaccuracy stemmed from the dependency on $\tau$, which needs to be either eliminated or added as a vector space within the classification to perform $\tau-invariant$ predictions.

\bibliography{mhi_bibliography}

\end{document}